\documentclass[10pt,twocolumn,letterpaper]{article}

\usepackage{ijcb}
\usepackage{times}
\usepackage{epsfig}
\usepackage{graphicx}
\usepackage{amsmath}
\usepackage{amssymb}
\usepackage{array,multirow}
\usepackage{threeparttable}
\usepackage{xcolor}
\usepackage{booktabs}
\usepackage{booktabs}
\usepackage{subfig}
\usepackage{multibib}

\newcites{SM}{References}

\newcommand*\rot{\rotatebox{90}}

\ijcbfinalcopy 

\ifijcbfinal\fi

\begin{document}

\title{Deception Detection and Remote Physiological Monitoring: A Dataset and Baseline Experimental Results}

\author{Jeremy Speth\thanks{Equal contribution.} , Nathan Vance\footnotemark[1] , Adam Czajka, Kevin W. Bowyer, Diane Wright, Patrick Flynn\\
University of Notre Dame\\
{\tt\small \{jspeth, nvance1, aczajka, kwb, dwright2, flynn\}@nd.edu}

}

\maketitle
\thispagestyle{empty}

\graphicspath{{./graphics/}}


\begin{abstract}
We present the Deception Detection and Physiological Monitoring (DDPM) dataset and initial
baseline results on this dataset.
Our application context is an interview scenario in which the interviewee attempts to deceive the interviewer on selected responses.
The interviewee is recorded in RGB, near-infrared, and long-wave infrared, along with cardiac pulse, blood oxygenation, and audio. After collection, data were
annotated for interviewer/interviewee, curated, ground-truthed, and organized into train / test parts for a set of canonical deception detection experiments.
Baseline experiments found random accuracy for micro-expressions as an indicator of deception, but 
that saccades can give a statistically significant response.
We also estimated subject heart rates from face videos (remotely) with a mean absolute error as low as 3.16 bpm.
The database contains almost 13 hours of recordings of 70 subjects, and over 8 million visible-light, near-infrared, and thermal video frames, along with appropriate meta, audio and pulse oximeter data. 
To our knowledge, this is the only collection offering recordings of five modalities in an interview scenario that can be used in both deception detection and remote photoplethysmography research.
\end{abstract}
\section{Introduction}

New digital sensors and algorithms offer the potential to address problems in human monitoring. Two interesting problems in this domain are remote physiological monitoring ({\em e.g.}, via remote photoplethysmography (rPPG)~\cite{Sun2016}) and deception detection (via remote analysis of various signals, such as pulse rate, blinking, or EEG that attempts to predict anxiety and/or cognitive load~\cite{zhang2007real, bhaskaran2011lie, vrij2000detecting,Duran2018,vrij2015saccadic, Gupta2019, Radlak2015,Rajoub2014,Owayjan2012}). In this paper, we present the Deception Detection and Physiological Monitoring (DDPM) dataset and baseline experiments with this dataset. DDPM is collected in an interview context, in which the interviewee attempts to deceive the interviewer with selected responses.
This task is motivated by, for example, applications in traveler screening.
DDPM supports analysis of video and pulse data for facial features including pulse, gaze, blinking, face temperature, and micro-expressions. The dataset comprises over 8 million high resolution RGB, NIR and thermal frames from face videos, along with cardiac pulse, blood oxygenation, audio, and deception-oriented interview data.
We provide this dataset with evaluation protocols to help researchers assess automated deception detection techniques.\footnote{https://cvrl.nd.edu/projects/data/\#deception-detection-and-physiological-monitoringddpm}
The {\bf main contributions} of this work are: (a) the largest {\bf deception detection dataset} in terms of total truthful and deceptive responses, recording length, and raw data size; (b) the first dataset for both deception detection and {\bf remote pulse monitoring} with RGB, NIR, and thermal imaging modalities; (c) the first rPPG dataset with {\bf facial movement and expressions} in a natural conversational setting; and (d) {\bf baseline results for deception detection} using non-visual saccadic eye movements, {\bf heart rate estimation} with five different approaches, and feature fusion results.

\section{Background}\label{sec:background}



\paragraph{Databases for deception detection.} Most research in deception detection  
has been designed and evaluated on private datasets, typically using a single sensing modality.
The DDPM dataset addresses these drawbacks. The top of Table~\ref{tab:datasets} compares sensor modalities and acquisition characteristics for existing datasets and DDPM.
Early researchers, inspired by the polygraph, believed information from the nervous system would likely give the best signals for deceit. Along these lines, the EEG-P300 dataset~\cite{Turnip2017} was proposed, which consists solely of EEG data. Early work by Ekman~\etal \cite{ekman2009lie} asserted that  humans could be trained to detect deception with high accuracy, using micro-expressions.
Inspired by human visual capabilities, the Silesian dataset~\cite{Radlak2015} contains
high frame-rate RGB video for more than 100 subjects. The Box-of-Lies dataset~\cite{Soldner2019} was released with RGB video and audio from a game show, and presents preliminary findings using linguistic, dialog, and visual features.
Multiple modalities have been introduced in the hope of enabling more robust detection. 
P\'{e}rez-Rosas \etal~\cite{Perez-Rosas2014} introduced a dataset for deception including RGB and thermal imaging, as well as physiological and audio recordings. DDPM includes these modalities but adds NIR imaging, higher temporal and spatial resolution in RGB, and twice as many interviews. Gupta \etal~\cite{Gupta2019} proposed Bag-of-Lies, a multimodal dataset with gaze data for detecting deception in casual settings.
Concerns about the authenticity of deception in constrained environments spurred the creation of the Real-life Trial dataset~\cite{Perez-Rosas2015}. Transcripts and video from the courtroom were obtained from public multimedia sources to construct nearly an hour of authentic deception footage. While the environment for ``high-stakes'' behavior is more difficult to achieve in the lab setting, the number of free variables involved in retrospectively assembling a real-world dataset (\eg, camera resolution, angle, lighting, distance) may make algorithm design difficult.
\paragraph{Databases for rPPG.} 
The first widely used publicly available dataset
was MAHNOB-HCI~\cite{Soleymani2012}, 
in which subjects' faces are relatively stationary (a significant limitation). 
Stricker \etal~\cite{Stricker2014} introduced PURE, the first public dataset with stationary and moving faces.
Later, MMSE-HR~\cite{Tulyakov2016} 
was used for rPPG during elicited emotion. The dataset consisted of more subjects than MAHNOB-HCI with more facial motion. The COHFACE dataset and open source implementations of three rPPG algorithms were introduced in~\cite{Heusch2017}. 
To accommodate data requirements for deep learning-based solutions, the VIPL-HR dataset~\cite{Niu2018} was created. Aside from being the largest publicly available dataset for rPPG, they released preliminary results from a CNN that outperformed then-existing techniques.
Recently, the 
UBFC-RPPG~\cite{Bobbia2019} dataset (containing rigid motion and a skin segmentation algorithm for rPPG) was released.
%
Table~\ref{tab:datasets} lists properties of these databases. To our knowledge, no rPPG dataset other than DDPM contains \textit{natural} conversational behavior with unconstrained facial movement.

\begin{table*}
    \caption{Comparison of the different modalities and environments for several databases for deception detection and rPPG.}
    \linespread{0.4}
    \begin{center}\vskip-5mm
    \begin{threeparttable}\footnotesize
    \begin{tabular}{ccccccccccccc}
        \toprule
        & \textbf{Dataset}
        & \begin{tabular}{@{}c@{}}\textbf{Subject}\\\textbf{Count}\end{tabular}
        & \begin{tabular}{@{}c@{}}\textbf{Length}\\\textbf{(Minutes)}\end{tabular}
        & \begin{tabular}{@{}c@{}}\textbf{Head}\\\textbf{Motion}\end{tabular}
        & \textbf{Talking} & \textbf{RGB} & \textbf{NIR} & \textbf{Thermal}
        & \begin{tabular}{@{}c@{}}\textbf{Physio-}\\\textbf{logical}\end{tabular}
        & \textbf{Audio}
        & \begin{tabular}{@{}c@{}}\textbf{Train/Test}\\\textbf{Splits}\end{tabular} & \begin{tabular}{@{}c@{}}\textbf{Raw}\\\textbf{Data}\end{tabular}\\
        \midrule
        & Silesian \cite{Radlak2015} & 101 & 186 & \checkmark & \checkmark & \checkmark &  &  &  &  &  & \\[1ex]
        & Multimodal \cite{Perez-Rosas2014} & 30 & - &  & \checkmark & \checkmark &  & \checkmark & \checkmark & \checkmark &  & -\\[1ex]
        & Real Trials \cite{Perez-Rosas2015} & 56 & 56 & \checkmark & \checkmark & \checkmark &  &  &  & \checkmark & & \\[1ex]
        & EEG-P300 \cite{Turnip2017} & 11 & - &  & \checkmark &  &  &  & \checkmark &  &  & \checkmark \\[1ex]
        & Box-of-Lies \cite{Soldner2019} & 26 & 144 & \checkmark & \checkmark & \checkmark &  &  &  &  \checkmark &  & \\[1ex]
        \rot{\rlap{~~~~~\textbf{Deception}}}
        & Bag-of-Lies \cite{Gupta2019} & 35 & \textless241 &  & \checkmark & \checkmark &  &  & \checkmark & \checkmark & \checkmark & \\[0.4ex]
        \midrule
        \\[-1ex] & MAHNOB-HCI \cite{Soleymani2012} & 27 & 264 &  &  & \checkmark &  &  & \checkmark &  &  &  \\[1ex]
        & PURE \cite{Stricker2014} & 10 & 60 & \checkmark & \checkmark & \checkmark &  &  & \checkmark & & & \checkmark\\[1ex]
        & MMSE-HR \cite{Tulyakov2016} & 40 & \textless102 & \checkmark &  & \checkmark &  & \checkmark & \checkmark &  &  & \\[1ex]
        & COHFACE \cite{Heusch2017} & 40 & 160 &  &  & \checkmark &  &  & \checkmark &  & \checkmark & \\[1ex]
        & VIPL-HR \cite{Niu2018} & 107 & 1150/380* & \checkmark & \checkmark & \checkmark & \checkmark &  & \checkmark &  & \checkmark & \\[1ex]
        \rot{\rlap{~~~~~~\textbf{rPPG}}}
        & UBFC-RPPG \cite{Bobbia2019} & 43 & 70 & \checkmark &  & \checkmark &  &  & \checkmark &  &  & \\[1ex]
        \midrule
        \\[-1.4ex] & \textbf{DDPM (Ours)} & \textbf{70} & \textbf{776} & \pmb{\checkmark} & \pmb{\checkmark} & \pmb{\checkmark} & \pmb{\checkmark} & \pmb{\checkmark} & \pmb{\checkmark} & \pmb{\checkmark} & \pmb{\checkmark} & \pmb{\checkmark}\\
        \bottomrule
    \end{tabular}
    \begin{tablenotes}\footnotesize
        \item[] -- Certain features could not be acquired. \textless The length is estimated using the maximum length of a recording described in literature. * Some NIR videos in VIPL-HR were discarded, so the length of the RGB/NIR videos are given individually.
        \end{tablenotes}
    \end{threeparttable}
    \end{center}
    \label{tab:datasets}
\end{table*}

\paragraph{Deception detection methods.}
Ekman \etal used his Facial Action Coding System (FACS)~\cite{ekman1997face} to detect \textit{microexpressions} and make inferences about deception~\cite{ekman2009lie}. Zhang \etal \cite{zhang2007real} leveraged FACS to detect deceitful 
facial expressions, which do not reproduce the full set of action units exhibited by a genuinely felt emotion. We investigate using facial expressions for the purpose of deception detection.
%
Bhaskaran \etal \cite{bhaskaran2011lie} leveraged eye movements to detect deceit. Psychology research has found that the frequency of non-visual saccades is doubled when a person is engaging long-term memory. Research using an fMRI device has further shown long-term memory engagement to be indicative of deceit~\cite{ganis2003neural}. A study comparing non-visual saccades between planned lies, spontaneous lies, and truth telling found that the eye movement rate (in saccades per second) was greater in 68\% of subjects when telling a spontaneous lie versus telling the truth~\cite{vrij2015saccadic}. We investigate these claims on the data collected in this study.
%
Caso \etal discovered that a deceptive person utilizes relatively fewer \textit{deictic} (pointing) gestures and more \textit{metaphoric} gestures (\eg, forming a fist as a metaphor for strength) when compared to a truthful person~\cite{caso2006impact}. Michael \etal take the subject's full body posture into account when detecting deceit~\cite{michael2010motion}.
Nonverbal clues to deceit remain a controversial topic~\cite{brennen2020research}. A meta analysis conducted by Luke \cite{luke2019lessons} suggests that all of the nonverbal clues to deceit existing in psychological literature could plausibly have arisen due to random chance as a result of small sample sizes and selective reporting. Our dataset enables researchers to more rigorously test claims regarding nonverbal clues to deceit as well as assertions to the contrary.

\section{Apparatus} \label{sec:apparatus}

\begin{figure*}[!htb]
    \captionsetup[subfloat]{labelformat=empty}
    \centering
    \subfloat{\includegraphics[height=0.14\textheight]{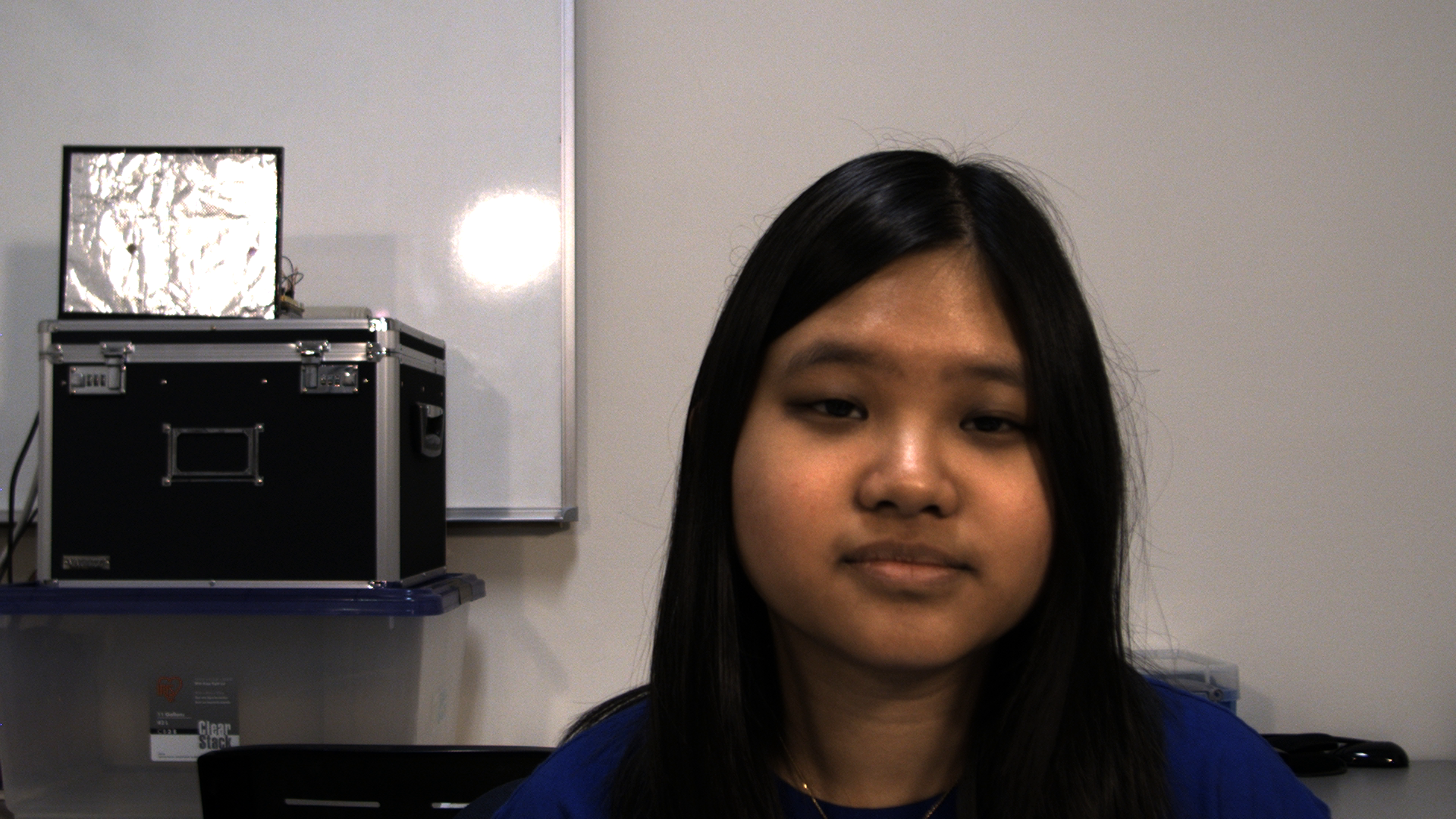}%
    \label{subfig_1}}
    \hfil
    \subfloat{\includegraphics[height=0.14\textheight]{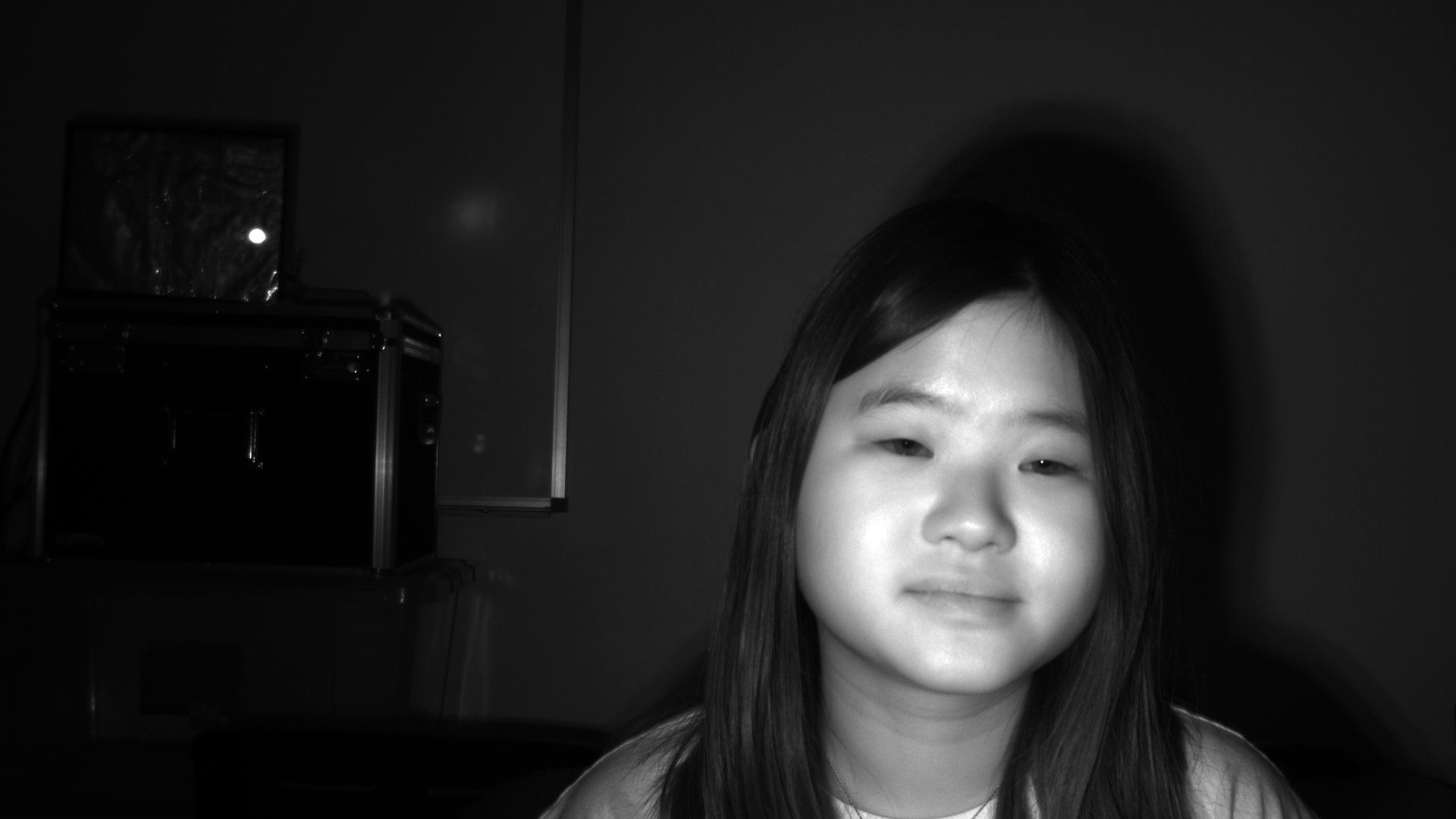}%
    \label{subfig_2}}
    \hfil
    \subfloat{\includegraphics[height=0.14\textheight]{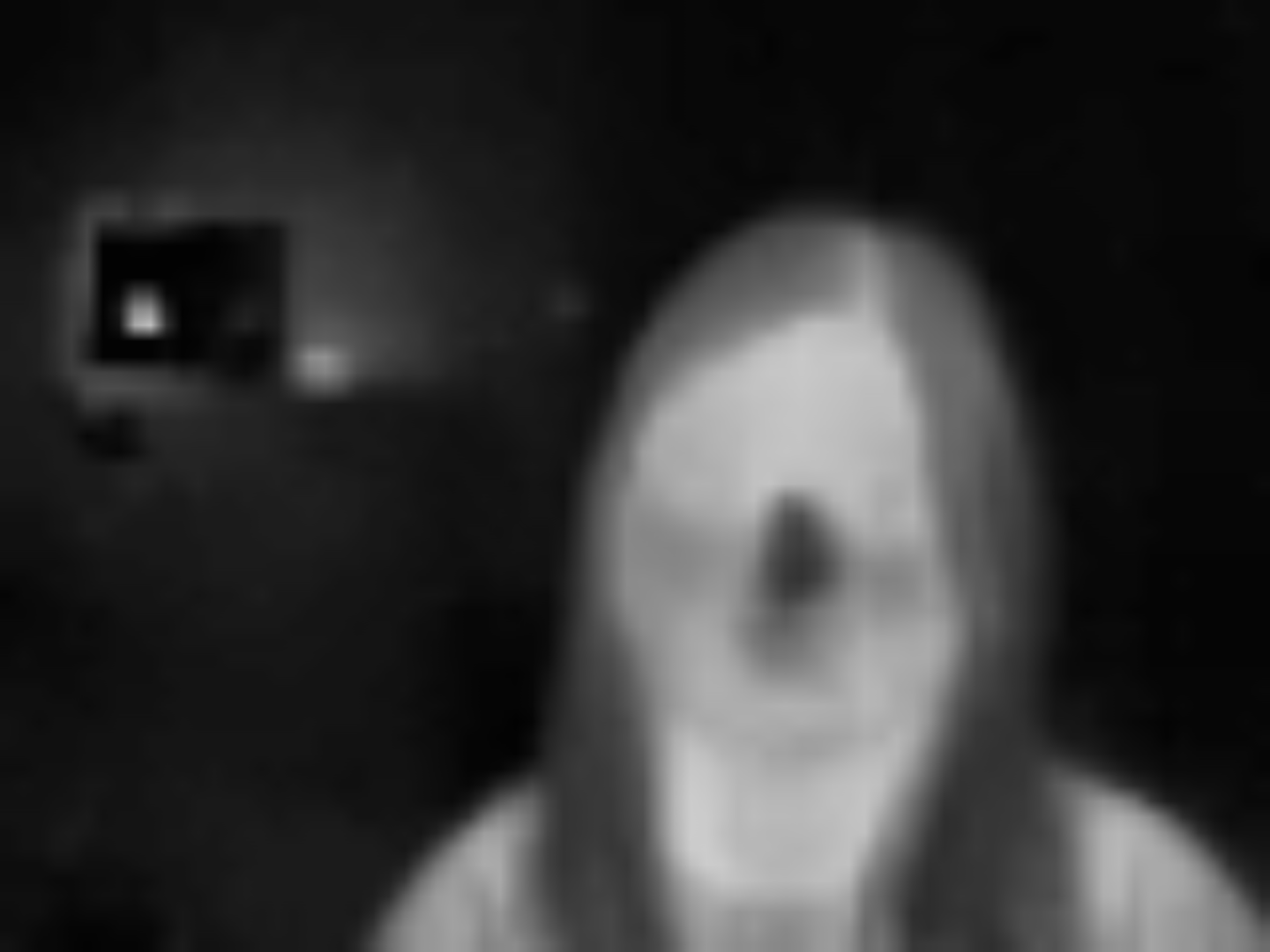}%
    \label{subfig_3}}
    \hfil
    \caption{Sample images from the RGB, NIR, and thermal cameras (left to right) from the collected DDPM dataset.
    }
    \label{fig:camera_samples}
\end{figure*}

\begin{figure}[!htb]
\centering
\includegraphics[width=0.44\textwidth]{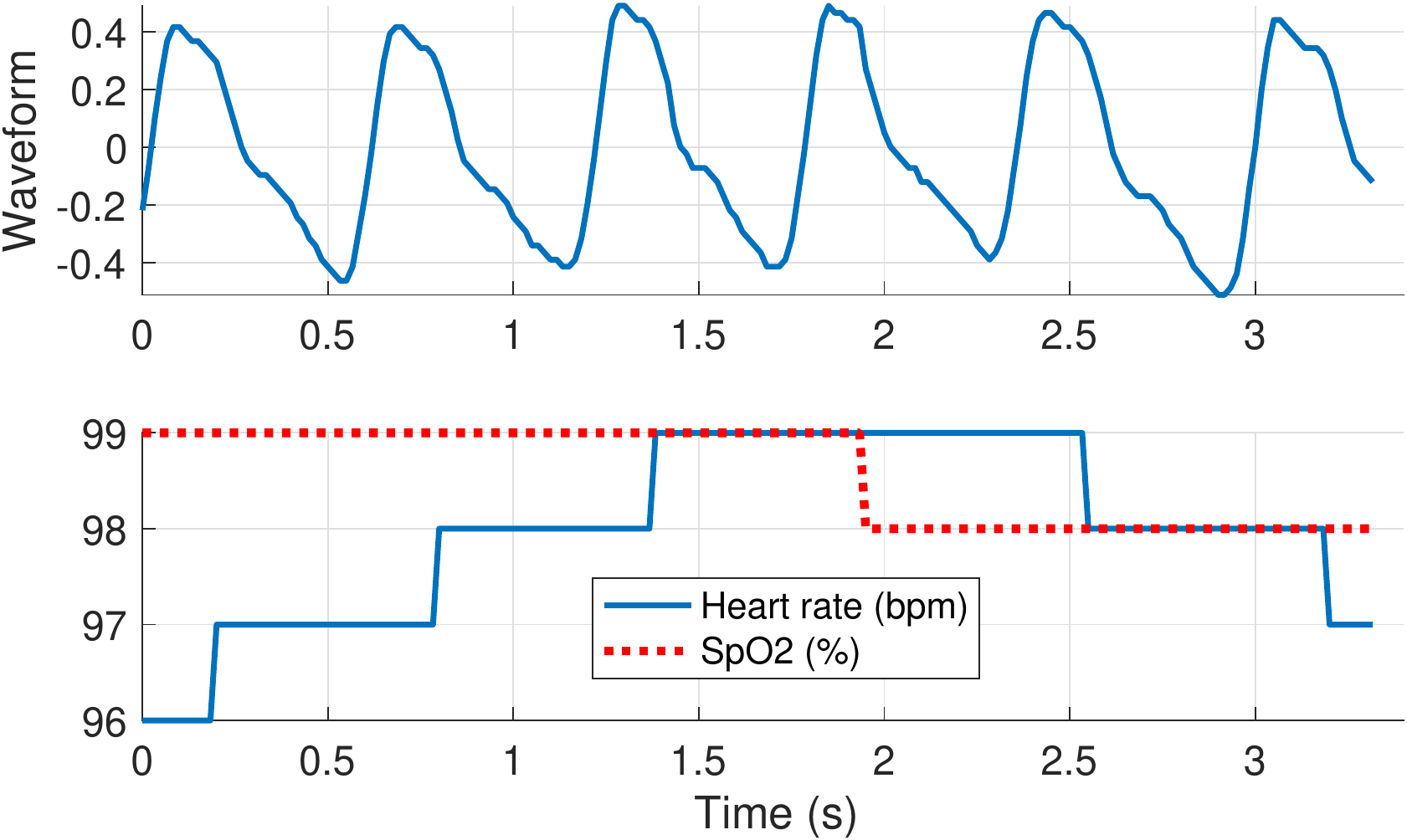}
\caption{Sample data recorded from the pulse oximeter, included into the collected DDPM dataset.}
\label{fig:oximeter_data}
\end{figure}

Detection of facial movements requires high spatial and temporal resolution. Analyzing images collected in different spectra as in Figure~\ref{fig:camera_samples} may provide deeper insight into facial cues associated with deception. Additionally, changes observed in the cardiac pulse rate as in Figure~\ref{fig:oximeter_data} may elucidate one's emotional state~\cite{Duran2018}. Speech dynamics such as tone changes provide another mode for detecting deception~\cite{Nasri2016}. We assembled an acquisition arrangement composed of three cameras, a pulse oximeter, and a microphone to address these needs. The sensing apparatus consisted of (i) a DFK 33UX290 RGB camera from The Imaging Source (TIS) operating at 90 FPS with a resolution of 1920 $\times$ 1080 px; (ii) A DMK 33UX290 monochrome camera from TIS with a bandpass filter to capture near-infrared images (730 to 1100 nm) at 90 FPS and 1920 $\times$ 1080 px; (iii) a FLIR C2 compact thermal camera that yielded 80 $\times$ 60 px images at 9 FPS; (iv) a FDA-certified Contec CMS50EA pulse oximeter that provides a 60 samples/second SpO2 and heart rate profile; and (v) a Jabra SPEAK 410 omni-directional microphone recording both interviewer and interviewee at 44.1 kHz with 16-bit audio measurements.  The sensors were time-synchronized using visible and audible artifacts generated by an Arduino-controlled device. Details of the synchronization design are provided in supplementary materials.  All data were captured by a workstation designed to accommodate the continuous streaming of data from the three cameras (750 Mbps), operating a GUI that contained subject registration and interview progression components.

\section{Procedure} \label{sec:procedure}

The collection protocol ensures that each subject gave both honest and deceptive answers.
Each session consisted of a brief preparatory meeting followed by a 10 - 20 minute interview in which 24 questions were asked, nine of which the subject was instructed to answer deceptively.  All data were collected under a protocol approved by the authors' institution's 
Human Subjects Institutional Review Board.

\paragraph{The mock interview setting.} A professional actor
was hired to conduct the interviews and to provide their judgment of truthful or deceptive answers to each question they asked.  The actor was instructed to be stoic and non-reactionary during the interviews, and to wear clothing consistent with that of a security official. They were supplied with an interview script to follow.
Subjects were motivated to deceive successfully through two levels of bonus compensation: if they were able to deceive the interviewer in five or six of the nine deceptive responses, they were given a 150 percent of a base incentive payment; the base payment was doubled if they were successfully deceptive in seven or more questions.


\paragraph{Interview content. }
Interview questions comprised three categories: experiential, travel screening, and ``superlative'' questions. The first group inquires whether or not the subject had a particular experience within the prior year, \eg  ``Have you traveled outside of the United States in the last 12 months?'' The second relates to an exercise conducted before the interview wherein the subject was given or asked to pack a suitcase with predetermined props, 
\eg ``Do you currently have any foreign currency in your possession?'' The final set of questions solicit the subject’s opinion on their best/worst or favorite/least favorite experience, \eg  ``What is your favorite flavor of ice cream?''
When answering such questions deceptively, subjects were to assert the opposite of the truth.
The first three ``warm up'' questions 
were always to be answered honestly. They allowed the subject to get settled,
and gave the interviewer an idea of the subject's demeanor when answering a question honestly. The order of the remaining questions and those selected for deception were randomly assigned for each subject.

Subjects were prepared for the interview by the Data Collection Coordinator (DCC). After signing a consent form, subjects were given a brief description of the recording equipment and an explanation 
of the experiment.
The categories of questions were described, but not specific questions, with the exception of the superlative questions that they were to answer deceptively.
The DCC emphasized that the more convincing the subject was vis-\`{a}-vis the interviewer, the more the subject would be compensated. At this point in interview preparation, either the subject or the DCC would pack the suitcase.
The subject was then given a survey to indicate which questions to answer deceptively and verify that they had answered the question according to the assignment.

\section{Collected Data} \label{sec:datasummary}

\definecolor{redrace}{RGB}{255,75,75}
\definecolor{redgender}{RGB}{185,50,50}
\definecolor{redage}{RGB}{125,25,25}
\definecolor{greenrace}{RGB}{75,255,75}
\definecolor{greengender}{RGB}{50,185,50}
\definecolor{greenage}{RGB}{25,125,25}
\definecolor{bluerace}{RGB}{75,75,255}
\definecolor{bluegender}{RGB}{50,50,185}
\definecolor{blueage}{RGB}{25,25,125}
\def\trainbar#1{{\color{red}\rule{#1mm}{5pt}}}
\def\trainbarrace#1{{\color{redrace}\rule{#1mm}{5pt}}}
\def\trainbargender#1{{\color{redgender}\rule{#1mm}{5pt}}}
\def\trainbarage#1{{\color{redage}\rule{#1mm}{5pt}}}
\def\valbarrace#1{{\color{greenrace}\rule{#1mm}{5pt}}}
\def\valbargender#1{{\color{greengender}\rule{#1mm}{5pt}}}
\def\valbarage#1{{\color{greenage}\rule{#1mm}{5pt}}}
\def\testbarrace#1{{\color{bluerace}\rule{#1mm}{5pt}}}
\def\testbargender#1{{\color{bluegender}\rule{#1mm}{5pt}}}
\def\testbarage#1{{\color{blueage}\rule{#1mm}{5pt}}}

\paragraph{Deception metadata.} Age, gender, ethnicity, and race were recorded for all participants. Each of the 70 interviews consisted of 24 responses, 9 of which were deceptive. Overall, we collected 630 deceptive and 1050 honest responses. To our knowledge, the 1,680 annotated responses is the most ever recorded in a deception detection dataset.

The interviewee recorded whether they had answered as instructed for each question.
For deceptive responses, they also rated how convincing they felt
they were, on a 5-point Likert scale ranging from ``I was not convincing at all'' to ``I was certainly convincing''. The interviewer recorded their belief about each response, on a 5-point scale from ``certainly the answer was deceptive'' to ``certainly the answer was honest''.
The data was additionally annotated to indicate which person (interviewer or interviewee) was speaking and the interval in time when they were speaking.

\paragraph{Data post-processing.} The RGB and NIR videos were losslessly compressed.
The interviews' average, minimum and maximum durations were 11 minutes, 8.9 minutes, and 19.9 minutes, respectively. In total, our dataset consists of 776 minutes of recording from all sensor modalities. The oximeter recorded SpO2, heart rate, and pulse waveform at 60 Hz giving average heart rates for the whole interview ranging from 40 bpm to 161 bpm. 

To encourage reproducible research, we defined subject-disjoint training, validation, and testing sets, with stratified random sampling across demographic features. Table \ref{tab:demographics} shows the demographics for each set. 

\newcommand*\rotbar{\rotatebox{270}}
\setlength\tabcolsep{2.8pt}
\begin{table}[!htb]\centering
    \caption{Number of subjects in various demographic categories across the training, validation, and test sets.}
    \label{tab:demographics}
    \footnotesize
    \begin{tabular}{@{}c cccccc c ccc c cccccc@{}}\toprule
    & \multicolumn{6}{c}{\textbf{Race}} && \multicolumn{3}{c}{\textbf{Gender}} && \multicolumn{6}{c}{\textbf{Age}}\\
    \cmidrule{2-7} \cmidrule{9-11} \cmidrule{13-18}
    & \rot{\textbf{White}}
        & \rot{\begin{tabular}{@{}c@{}}\textbf{African}\\[-0.6ex]\textbf{American}\end{tabular}}
        & \rot{\textbf{Chinese}}
        & \rot{\begin{tabular}{@{}c@{}}\textbf{Asian}\\[-0.6ex]\textbf{Indian}\end{tabular}} & \rot{\textbf{Filipino}} & \rot{\textbf{Other}}
        && \rot{\textbf{Female}} & \rot{\textbf{Male}} & \rot{\textbf{Nonbinary}}
        && \rot{\textbf{18-19}} & \rot{\textbf{20-29}} & \rot{\textbf{30-39}} & \rot{\textbf{40-49}} & \rot{\textbf{50-59}} & \rot{\textbf{60+}}\\
        \cmidrule{2-7} \cmidrule{9-11} \cmidrule{13-18}
        
        
        
        Train & 33 & 3 & 3 & 3 & 1 & 5 && 31 & 16 & 1 && 4 & 32 & 2 & 4 & 5 & 1\\ 
        Val. & 6 & 1 & 2 & 0 & 1 & 1 && 6 & 5 & 0 && 0 & 9 & 1 & 0 & 1 & 0 \\
        Test & 6 & 1 & 1 & 1 & 0 & 2 && 7 & 4 & 0 && 0 & 7 & 2 & 1 & 1 & 0\\
        \bottomrule
    \end{tabular}
\end{table}
\section{Pulse Detection Experiments} \label{sec:pulse}

Five pulse detection techniques were evaluated on the DDPM dataset,
relying on {\bf blind-source separation}~\cite{Poh2010, Poh2011}, {\bf chrominance and color space transformations}~\cite{DeHaan2013, Wang2017}, and {\bf deep learning}~\cite{Yu2019}. All methods are the authors' implementations based on the published descriptions.

The general pipeline for pulse detection contains region selection, spatial averaging, a transformation or signal decomposition, and frequency analysis. For region selection, we used OpenFace~\cite{baltrusaitis2018openface} to detect 68 facial landmarks used to define a face bounding box.
The bounding box was extended horizontally by 5\% on each side, and by 30\% above and 5\% below, and then converted to a square with a side length that was the larger of the expanded horizontal and vertical sizes, to ensure that the cheeks, forehead and jaw were contained.
For the chrominance-based approaches, we select the skin pixels within the face with the method of Heusch \etal~\cite{Heusch2017}.

Given the region of interest, we performed channel-wise spatial averaging to produce a 1D temporal signal for each channel. The blind source separation approaches apply ICA to the channels, while the chrominance-based approaches combine the channels to define a robust pulse signal. The heart rate is then found over a time window by converting the signal to the frequency domain and selecting the peak frequency $f_p$ as the cardiac pulse. The heart rate is computed as $\widehat{HR} = 60 \times f_p$ beats per minute (bpm).

For training the deep learning-based approach, we employed 3D Convolutional Neural Network (3DCNN), fed with the face cropped to bounding box and downsized to 64$\times$64 with bicubic interpolation. During training and evaluation, the model is given clips of the video consisting of 135 frames (1.5 seconds). We selected this as the minimum length of time an entire heartbeat would occur, considering 40 beats per minute (bpm) as a lower bound for average subjects. The 3DCNN was trained to minimize the negative Pearson correlation between predicted and normalized ground truth pulse waveforms. 

The oximeter recorded ground truth waveform and heart rate estimates at 60 Hz
and upsampled to 90 Hz to match the RGB camera frame rate.
One of the difficulties in defining an oximeter waveform as a target arises from the phase difference observed at the face and finger and time lags from the acquisition apparatus \cite{Zhan2020}.
To mitigate the phase shift, we use the output waveform predicted by CHROM to shift the ground truth waveform, such that the cross-correlation between them is maximized. We use the Adam optimizer with a learning rate of $\alpha=0.0001$, and parameter values of $\beta_1=0.99$ and $\beta_2=0.999$ to train the model for 50 epochs, then select the model with the lowest loss on the validation set as our final model.

For videos longer than the clip length of 135 frames it is necessary to perform predictions in sliding window fashion over the full video. Similar to~\cite{DeHaan2013}, we use a stride of half the clip length to slide across the  video. The windowed outputs are standardized, a Hann function is applied to mitigate edge effects from convolution, and they are added together to produce a single value per frame.

Pulse detection performance is analyzed by calculating the error between heart rates for time periods
of length 30 seconds with stride of a single frame. We apply a Hamming window prior to converting the signal to the frequency domain and select the index of the maximum spectral peak between between $0.6\overline{6}$ Hz and 3 Hz (40 bpm to 180 bpm) as the heart rate. A five-second moving-window average filter is then applied to the resultant heart rate signal to smooth noisy regions containing finger movement. We used metrics from the rPPG literature to evaluate performance, such as mean error (ME), mean absolute error (MAE), root mean squared error (RMSE), and Pearson correlation coefficient of the heart rate, $r$, as shown in Tab.~\ref{tab:rppg_results}. The original blind-source separation approach, POH10, is outperformed by POH11 due to signal detrending and filtering, which removes noise from motion. Both chrominance-based approaches perform similarly, although POS gives good accuracy without filtering. The 3DCNN gives results slightly worse than the chrominance-based methods, yet still accurate and deserving attention.
%

\section{Non-visual Saccades Experiments} \label{sec:saccades}

Saccades are rapid eye movements between points of fixation. Visual saccades are common when a person shifts attention between objects. However, not all saccades occur for visual purposes. 
In a study comparing \textit{non-visual saccades} (which occur without a vision-related purpose) between planned lies, spontaneous lies, and truth telling, psychology researchers found that the Eye Movement Rate (EMR, in saccades per second) was greater in 68\% of subjects when telling a spontaneous lie versus telling the truth, with no statistical difference between planned and spontaneous lies, nor between planned lies and truth~\cite{vrij2015saccadic}. 

\definecolor{darkgreen}{RGB}{0,150,0}
\begin{table}[!htb]
    \begin{center}
    \caption{Comparison between pulse estimators across \textbf{DDPM}, \textcolor{red}{MAHNOB-HCI}\cite{Soleymani2012}, \textcolor{darkgreen}{VIPL-HR}\cite{Niu2018}, and \textcolor{blue}{UBFC-RPPG}\cite{Bobbia2019}. {\footnotesize \textcolor{red}{*} PhysNet-3DCNN-ED \cite{Yu2019} (slightly different than 3DCNN)}
    }
    \label{tab:rppg_results}
    \resizebox{\columnwidth}{!}{%
    \begin{tabular}{ccccc}
        \toprule
        Method & \begin{tabular}{@{}c@{}}ME \\ (bpm)\end{tabular} & \begin{tabular}{@{}c@{}}MAE \\ (bpm)\end{tabular} & \begin{tabular}{@{}c@{}}RMSE \\ (bpm)\end{tabular} & $r$\\
        \midrule
        CHROM~\cite{DeHaan2013} & \bf{-0.26} & \bf{3.48} & \textbf{10.4} / \textcolor{red}{10.7}                                / \textcolor{darkgreen}{16.9} / \textcolor{blue}{2.39}
                                & \textbf{0.93} / \textcolor{red}{0.82\phantom{*}}                     / \textcolor{darkgreen}{0.28} / \textcolor{blue}{0.96}\\
        POS~\cite{Wang2017}   & \bf{0.11} & \bf{3.16}  & \textbf{11.2} / \phantom{x}\textcolor{red}{-}\phantom{.x}            / \textcolor{darkgreen}{17.2} / \textcolor{blue}{6.77}
                                & \textbf{0.92} / \phantom{x}\textcolor{red}{-}\phantom{.x\phantom{*}} / \textcolor{darkgreen}{0.30} / \textcolor{blue}{0.96}\\
        POH10~\cite{Poh2010}   & \bf{18.54}         & \bf{20.56} & \textbf{33.1} / \textcolor{red}{25.9}            / \phantom{x}\textcolor{darkgreen}{-}\phantom{.x} / \phantom{x}\textcolor{blue}{-}\phantom{.x}
                                & \textbf{0.56} / \textcolor{red}{0.08\phantom{*}} / \phantom{x}\textcolor{darkgreen}{-}\phantom{.x} / \phantom{x}\textcolor{blue}{-}\phantom{.x}\\
        POH11~\cite{Poh2011}    & \bf{10.47} & \bf{14.30} & \textbf{28.9} / \textcolor{red}{13.6}            / \phantom{x}\textcolor{darkgreen}{-}\phantom{.x} / \phantom{x}\textcolor{blue}{-}\phantom{.x}
                                & \textbf{0.54} / \textcolor{red}{0.36\phantom{*}} / \phantom{x}\textcolor{darkgreen}{-}\phantom{.x} / \phantom{x}\textcolor{blue}{-}\phantom{.x}\\
        3DCNN~\cite{Yu2019}     & \bf{0.54}  & \bf{4.11} & \textbf{11.9} / \textcolor{red}{7.9*}            / \phantom{x}\textcolor{darkgreen}{-}\phantom{.x} / \phantom{x}\textcolor{blue}{-}\phantom{.x}
                                & \textbf{0.92} / \textcolor{red}{0.76*}           / \phantom{x}\textcolor{darkgreen}{-}\phantom{.x} / \phantom{x}\textcolor{blue}{-}\phantom{.x}\\
        \bottomrule
    \end{tabular}%
    }
    \end{center}
\end{table}

\begin{table}[!htb]
    \centering
    \caption{Comparison of Eye Movement Rate for eye saccades measured for truthful and deceitful answers with a paired-sample t-test}
    \footnotesize
    \begin{tabular}{cccc}
        \toprule
        Study & \begin{tabular}{@{}c@{}}Average\\ difference (EMR)\end{tabular} & \begin{tabular}{@{}c@{}}\% subjects\\ follow trend\end{tabular} & \textit{p}-value \\
        \midrule
        Vrij \textit{et al.}~\cite{vrij2015saccadic} & 0.13 & 68\% & \textbf{0.013} \\
        Ours (Combined) & -0.00034 & 52\% & 0.585 \\
        Ours (Question) & 0.00048 & 44\% & 0.687 \\
        Ours (Response) & 0.0031 & 64\% & \textbf{0.0098} \\
        \bottomrule
    \end{tabular}
    \label{tab:saccades}
\end{table}

We calculate the 
EMR
using a 3-step process: gaze angle extraction, denoising, and saccade identification. 
We extract the eye gaze angle using OpenFace~\cite{baltrusaitis2018openface}. 
While the OpenFace gaze tracker is sensitive to minute eye movements, it also can be prone to noise. We found that, in addition to averaging the gaze angle between the two eyes, it was necessary to average across every 3 frames as well. For every pair of frames, we calculate the angular velocity of the eye using both the vertical and horizontal rotation of the eyes. If the rotational velocity exceeds a threshold (set experimentally to 50 degrees per second), then a saccade has occurred. The average, standard deviation, min and max durations (in frames) for the 86,168 detected saccades are 3.50, 2.92, 1 and 10, respectively.





Non-visual saccade research provides a hypothesis for us to test: subjects telling a spontaneous lie have a statistically significant increase in saccade rate over subjects telling the truth~\cite{vrij2015saccadic}. 
We compared the eye movement between truthful and deceptive responses using a pairwise t-test, where the pairings were the average truthful versus deceptive EMR for each subject. We compared the EMR for response portions since this is the time period in which the subject is actively telling a lie, and also for the question portions during which the subject is not yet actively telling a lie. 
We found that subjects exhibit a statistically significant increase in EMR when giving a deceptive response, with 64\% of subjects following that trend; see Table~\ref{tab:saccades}. We did not find a significant change in EMR for time intervals while the subject is hearing the question, nor the union of the question and response intervals. These results confirm the findings of Vrij \etal~\cite{vrij2015saccadic}. When used in a simple thresholding classifier, which partitions truthful and deceptive responses on each subject's median saccade rate, we obtain a discrimination accuracy of 54.5\%.
\section{Pulse and Saccades Fusion}

\begin{figure}[!htb]
\centering
\includegraphics[width=0.8\linewidth]{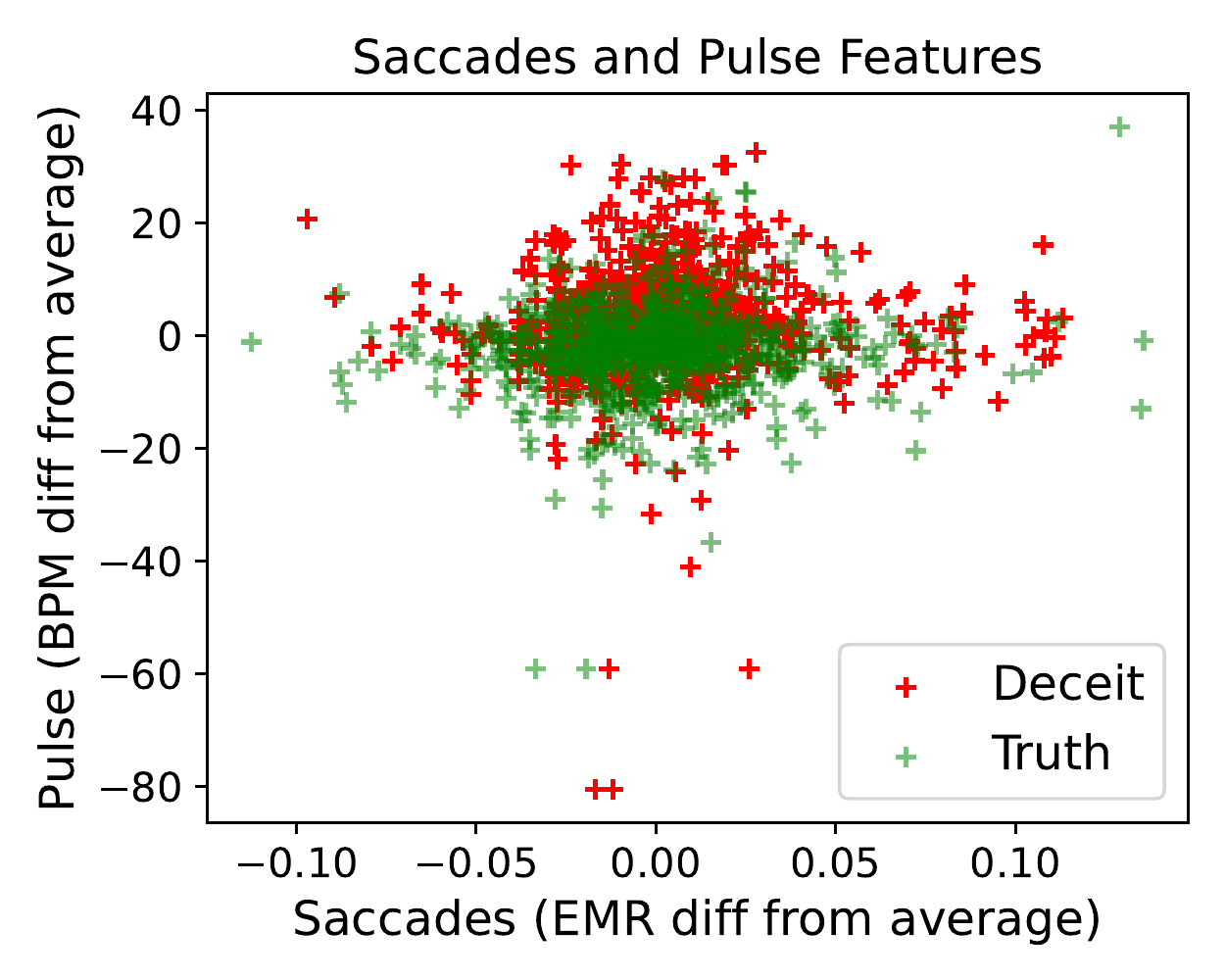}
\caption{Separability of deceptive and truthful answers by pulse and saccades features.}
\label{fig:featurefusion}
\end{figure}

We attempted to combine our two promising features, pulse and saccades, for the purpose of deception detection. We used the same simple thresholding classifier as in Section \ref{sec:saccades}. Classification results are shown in Table~\ref{tab:classifier}.

\begin{table}[!htb]
    \centering
    \caption{Deception detection accuracy by technique.}
    \footnotesize
    \begin{tabular}{cc|cc}
        \toprule
        Strategy & Accuracy & Strategy & Accuracy \\
        \midrule
        Pulse & 63.4\% & Saccades & 54.5\% \\
        Pulse \textit{OR} Saccades & 59.8\% & Pulse \textit{AND} Saccades & 58.1\% \\
        SVM (Linear) & 62.1\% & SVM (RBF) & 62.6\% \\
        \bottomrule
    \end{tabular}
    \label{tab:classifier}
\end{table}

We compare the classification accuracy for pulse and saccades as used on their own, and several simple fusion techniques. The \textit{OR} and \textit{AND} techniques are to ``or" or ``and" the results of the individual classifiers. The SVM techniques utilize the sklearn SVC package. For train and test splits, we use the splits published with the DDPM dataset, merging the train and validate splits for training over 59 subjects (1239 questions) and testing over 11 subjects (231 questions).

We found that pulse and saccadic signals are not orthogonal as shown in Figure~\ref{fig:featurefusion}, with their combination being unable to outperform pulse on its own. Because the pulse signal is more reliable, \ie only exposed skin is required as opposed to a direct line of sight to the subject's eyes in the case of saccades, we recommend that deception detection researchers prioritize the pulse signal.
\section{Microexpressions}\label{sec:micro}

Microepressions are defined as brief, involuntary facial expressions lasting less than half a second~\cite{ekman1969nonverbal}.
Their presence is believed by some in Psychology to be indicative of ongoing deceit~\cite{ekman2009telling}.


Inspired by the work of Yap \etal~\cite{yap2020samm}, we hypothesize that when a microexpression occurs we will see a deviation from the at-rest expression, measured in facial action units (FAUs). We acknowledge that, while this matches closely with the definition of microexpressions, it may not always be the case when the subject has many manipulators (\ie conscious facial motions) or is speaking. 
%
%
Our method for spotting microexpressions consists of the following steps: extract all 18 facial action unit (FAU) features offered by OpenFace~\cite{baltrusaitis2018openface}, calculate the likelihood, onset time, and offset time of potential microexpressions, and select non-overlapping candidates with likelihoods above a threshold.
The model of a microexpression is an increase followed by a decrease in FAU signals, in a short window of time. Therefore, we detect microexpressions using a sliding window where the start of the window corresponds to the onset of the microexpression, the center to the apex, and the end to the offset. The likelihood is proportional to the difference of the apex to the onset and offset. We use multiple fixed-length windows to account for variable-length microexpressions. The threshold to which the likelihood values are compared is dynamic and calculated based on the subject's baseline response.
Parameters of our microexpression detection model were set experimentally with two published microexpression datasets, CAS(ME)$^2$~\cite{qu2017cas} and SAMM~\cite{davison2016samm}.
Because we do not have microexpression ground truth in the proposed DDPM dataset, 
we develop our model by performing parameter optimization on one of the two microexpression datasets and evaluating on the other. This strategy increases our confidence that the method may be transferred between datasets. Using this cross-dataset evaluation, we obtained F1 scores of 0.0264 and 0.0509 on CAS(ME)$^2$ and SAMM, respectively, as shown in Tab.~\ref{tab:precision-recall}.

\begin{table}[!ht]
    \centering
    \caption{Cross-dataset evaluation results in microexpression detection experiments.}
    \footnotesize
    \begin{tabular}{lccc}
        \toprule
        & Precision & Recall & $F_1$ \\ 
        \midrule
        CAS(ME)$^2$ & 0.0136 & 0.439 & 0.0264  \\
        SAMM & 0.0349 & 0.0943 & 0.0509\\
        \bottomrule
    \end{tabular}
    \label{tab:precision-recall}
\end{table}
\vskip-2mm

We compare the $F_1$ scores that we obtained to the $F_1$ scores of several state of the art microexpression spotting techniques in Table~\ref{tab:baseline-comparison}. The state of the art techniques arose from the recent microexpression spotting competition, the Micro-Expression Grand Challenge 2020~\cite{li2020megc2020}. We found that our model is comparable to the existing techniques, being beaten only by the optical flow based technique proposed by Zhang et al.~\cite{zhang2020fusion}. The $F_1$ scores reported by Pan are averaged scores for both microexpression and macroexpression spotting and thus should not be compared directly with the other methods~\cite{pan2020local}.

\begin{table}[!ht]
    \centering
    \caption{$F_1$ scores for microexpression spotting on CAS(ME)$^2$ and SAMM datasets}
    \footnotesize
    \begin{tabular}{ccc}
        \toprule
        Method & CAS(ME)$^2$ & SAMM Long Videos \\
        \midrule
        Baseline~\cite{he2019spotting} & 0.0082 & 0.0364 \\
        *Pan~\cite{pan2020local} & 0.0595 & 0.0813 \\
        Zhang et al.~\cite{zhang2020fusion} & 0.0547 & 0.1331 \\
        Yap et al.~\cite{yap2020samm} & - & 0.0508 \\
        Ours & 0.0264 & 0.0509 \\
        \bottomrule
    \end{tabular}
    \label{tab:baseline-comparison}
\end{table}


We applied our microexpression extraction method to the DDPM dataset, obtaining intervals in which microexpressions are predicted to occur. We conducted a paired sample t-test comparing microexpression rates between truthful and deceptive responses for each subject. Because we did not know whether microexpressions should be more prevalent during the question asking phase or the answering phase, we conducted the test under three conditions: comparing microexpression rates over the entire question-response time window, over just the question, and over just the response. The results are given in Tab.~\ref{tab:microexpressions}.
We did not find microexpressions as detected by state of the art techniques to be indicative of deceit. 

\begin{table}[!htb]
    \centering
    \caption{Comparison of microexpressions with a paired-sample t-test. Average difference is between truthful and deceptive in units of Microexpressions per Second.}
    \footnotesize
    \begin{tabular}{cccc}
        \toprule
        \begin{tabular}{@{}c@{}}Question\\ part\end{tabular} & \begin{tabular}{@{}c@{}}Average\\ difference\end{tabular} & \begin{tabular}{@{}c@{}}\% subjects\\ follow trend\end{tabular} & \textit{p}-value \\
        \midrule
        Combined & 2.74e-5 & 51\% & 0.583 \\
        Question only & -6.30e-5 & 50\% & 0.618 \\
        Response only & 7.797e-5 & 49\% & 0.666 \\
        \bottomrule
    \end{tabular}
    \label{tab:microexpressions}
\end{table}


\section{Conclusions} \label{sec:conclusion}

We present the Deception Detection and Physiological Monitoring (DDPM) dataset, the most comprehensive dataset to date in terms of number of different modalities and volume of raw video, to support exploration of deception detection and remote physiological monitoring in a natural conversation setup. The sensors are temporally synchronized, and imaging across visible, NIR and LWIR spectra provides more than 8 million high-resolution images from almost 13 hours of recordings in a deception-focused interview scenario.

Along with this dataset, we provide baseline results for heart rate detection, 
and the feasibility of deception detection using microexpressions, non-visual saccades, and heart rate. Non-visual eye saccades and heart rate to classify responses as deceptive gives statistically significant results. In contrast, we did not find microexpressions to be a reliable signal to deceit. As microexpression detectors become more accurate they may become a viable feature for deception detection.
This new dataset and baseline approaches are made publicly available with the mentioned evaluation protocols to further advance research in the areas of remote physiological monitoring and deception detection.
\paragraph{Acknowledgements.} We would like to thank Marybeth Saunders for conducting the interviews during data collection. This research was sponsored by the Securiport Global Innovation Cell, a division of Securiport LLC. Commercial equipment is identified in this work in order to adequately specify or describe the subject matter. In no case does such identification imply recommendation or endorsement by Securiport LLC, nor does it imply that the equipment identified is necessarily the best available for this purpose.  The opinions, findings, and conclusions or recommendations expressed in this publication are those of the authors and do not necessarily reflect the views of our sponsors.

{\small
\bibliographystyle{ieee}
\bibliography{main}
}

\clearpage
\appendix
\renewcommand\thefigure{\thesection.\arabic{figure}}
\renewcommand\thetable{\thesection.\arabic{table}}
\setcounter{figure}{0}
\setcounter{table}{0} 
\section*{Supplementary Material}

\section{Synchronization Device}

\begin{figure}[!htb]
\includegraphics[width=0.48\textwidth]{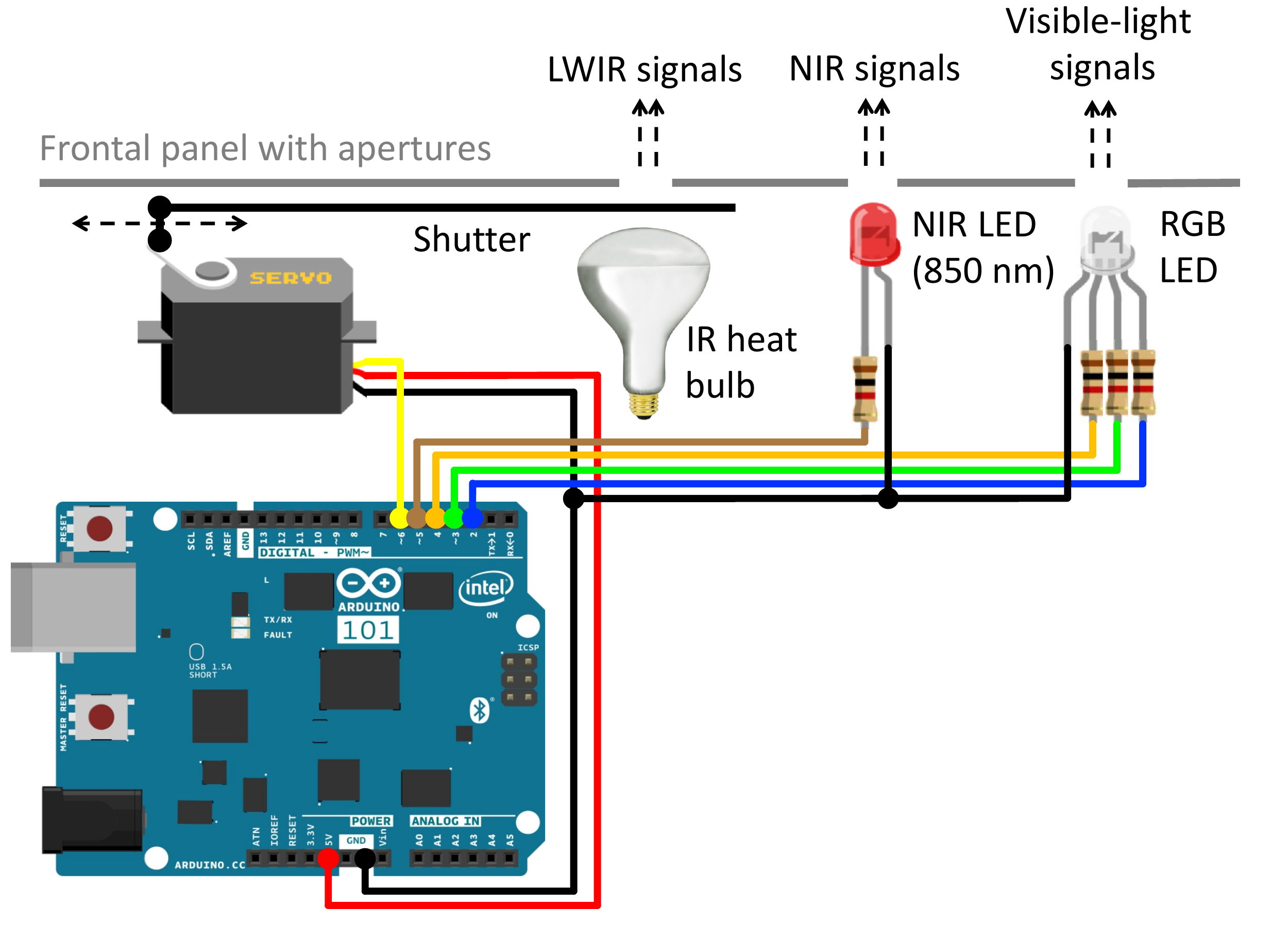}
\caption{Synchronization target generating signals in visible, near infrared and thermal spectra.}
\label{fig:Frankie}
\end{figure}

\begin{figure}[!htb]
\includegraphics[width=0.46\textwidth]{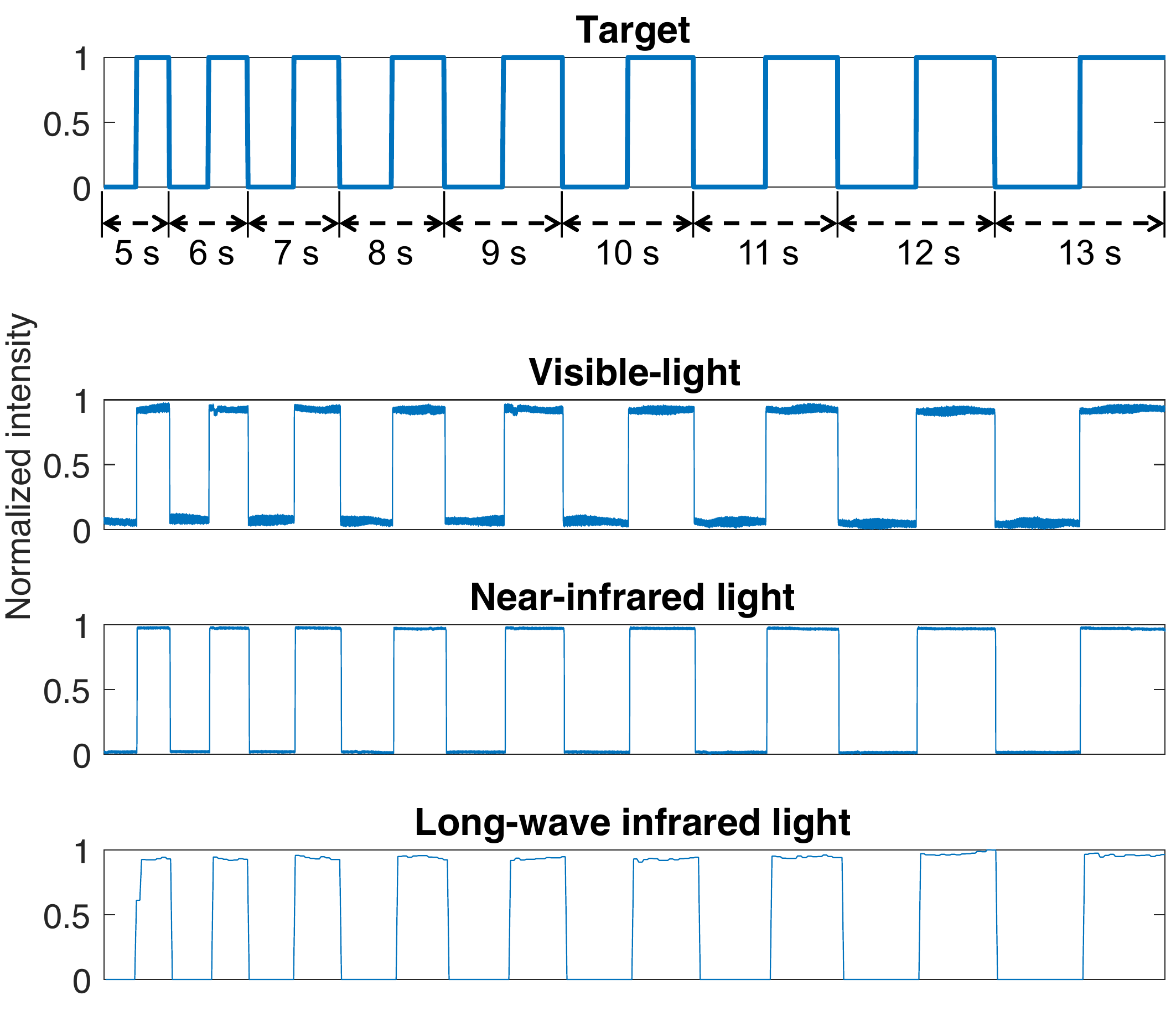}
\caption{Synchronization signal generated by the target (top) and its example reconstructions done by three sensors used in this study. The reconstructed intensities were normalized to $\langle 0,1 \rangle$ as the absolute signal value does not play a role in synchronization.}
\label{fig:Signals}
\end{figure}

Fig.~\ref{fig:Frankie} illustrates the synchronization device, based on Arduino board, that generates signals in visible, near infrared and thermal spectra with pre-defined duty cycle. While the duty cycle is kept constant (50\% of the period), we gradually increase the period from 5 seconds to 13 seconds, as shown in Fig.~\ref{fig:Signals}, to create a pattern that repeats every 81 seconds. Signals modulated in this way allow for cameras to be unambiguously synchronized over longer synchronization times, \ie up to 81 seconds, with multiple signal edges in between for finer alignment. In the current setup, a 50 ms delay is added to RGB and NIR signals due to inertia of the shutter used to generate the LWIR signal. Assuming that the position of the synchronization target is known within the video frame, the signal reconstruction is straightforward and based on reading local image intensity, as shown in Fig.~\ref{fig:Signals}.

\end{document}